\documentclass[letterpaper, 10 pt, conference]{ieeeconf}  

\IEEEoverridecommandlockouts                              

\usepackage{amsthm}
\usepackage{amsmath}
\usepackage{xcolor}
\usepackage{comment}
\usepackage{paralist}
\usepackage{hyperref}
\usepackage{graphicx}
\usepackage{amsfonts}
\usepackage[english]{babel}
\usepackage{booktabs}
\usepackage{multirow}
\usepackage{subcaption}
\usepackage{floatrow}
\usepackage{sidecap}
\usepackage[affil-it]{authblk}
\usepackage[T1]{fontenc}
\usepackage[scientific-notation=true]{siunitx}
\usepackage{mdframed}
\usepackage{makecell}

\usepackage[normalem]{ulem}
\usepackage{mathtools}
\usepackage{upgreek}

\DeclarePairedDelimiter{\norm}{\lVert}{\rVert}
\theoremstyle{definition}
\newtheorem{definition}{Definition}[section]

\title{Gradient-based Regularization for Action Smoothness in Robotic Control with Reinforcement Learning}

\author[1]{I Lee$^*$}
\author[1]{Hoang-Giang Cao$^*$}
\author[1]{Cong-Tinh Dao}
\author[1]{Yu-Cheng Chen}
\author[1,2]{I-Chen Wu$^\dagger$}

\affil[1]{Department of Computer Science, National Yang Ming Chiao Tung University, Taiwan}
\affil[2]{Research Center for IT Innovation, Academia Sinica, Taiwan}

\begin{document}
\maketitle

\def\thefootnote{$*$}\footnotetext{Equal contribution.}\def\thefootnote{\arabic{footnote}}
\def\thefootnote{$\dagger$}\footnotetext{Correspondence.}\def\thefootnote{\arabic{footnote}}

\thispagestyle{empty}
\pagestyle{empty}


\begin{abstract}
Deep Reinforcement Learning (DRL) has achieved remarkable success, ranging from complex computer games to real-world applications, showing the potential for intelligent agents capable of learning in dynamic environments.
However, its application in real-world scenarios presents challenges, including the jerky problem, in which jerky trajectories not only compromise system safety but also increase power consumption and shorten the service life of robotic and autonomous systems.
To address jerky actions, a method called conditioning for action policy smoothness (CAPS) was proposed by adding regularization terms to reduce the action changes.
This paper further proposes a novel method, named Gradient-based CAPS (Grad-CAPS), that modifies CAPS by reducing the difference in the gradient of action and then uses displacement normalization to enable the agent to adapt to invariant action scales. 
Consequently, our method effectively reduces zigzagging action sequences while enhancing policy expressiveness and the adaptability of our method across diverse scenarios and environments.
In the experiments, we integrated Grad-CAPS with different reinforcement learning algorithms and evaluated its performance on various robotic-related tasks in DeepMind Control Suite and OpenAI Gym environments.
The results demonstrate that Grad-CAPS effectively improves performance while maintaining a comparable level of smoothness compared to CAPS and Vanilla agents.

\end{abstract}


\section{Introduction}

In recent years, deep reinforcement learning (DRL) has achieved numerous milestones in many challenging computer games, ranging from traditional board games like Go, as demonstrated by AlphaGo \cite{silver2016alphago}, to complicated real-time strategy (RTS) video games such as Starcraft with AlphaStar \cite{Vinyals2019Grandmaster} and Dota2 with OpenAI Five \cite{openai2019dota}.
DRL has also been applied to various real-world control tasks, such as robotics and autonomous systems.
This holds the promise of intelligent agents capable of learning to make decisions through interactions within complex and dynamic environments.

However, applying DRL to real-world applications presents some challenges, particularly when it results in a jerky trajectory (\cite{siddharth2021caps,kobayashi2022l2c2,lipsnet_2023,cao2023imagebased}).
During the training process, DRL primarily focuses on maximizing accumulated rewards, typically without enforcing constraints related to trajectory smoothness.
Thus, lack of trajectory smoothness gives rise to critical issues, especially in autonomous systems such as robot manipulation or autonomous driving, where unpredictable behaviors are likely to be dangerous.
Such unstable control not only compromises the safety of the system but also significantly increases power consumption, resulting in mechanical wear and tear, reducing energy efficiency, and, consequently, the service life of the robot or vehicle.
Tackling the problem of jerky behaviors emphasizes the importance of stability and control issues in implementing DRL in real-world scenarios.
This effort enhances system performance and ensures the safe, reliable, and sustainable operation of robotic and autonomous systems across various applications. 

To address the jerky problem, a straightforward approach is to use reward engineering. 
Reward engineering is based on prior human knowledge about the tasks to design a specific reward function to penalize unsmooth trajectory (\cite{Hwangbo2017Control, Koch2019Flight, carlucho201871}). 
However, this approach is task-specific, limiting its generalizability to apply to other tasks.

Another approach is to use DRL with a hierarchical network structure. 
The objective is to optimize the overall episode reward while simultaneously mitigating control or action oscillations, as demonstrated in the works of Yu et al. (2021) \cite{Yu2021TAAC} and Chen et al. (2021) \cite{Chen2021NestedSoftAC}.
Mysore {\em et al.} proposed conditioning for action policy smoothness (CAPS) for solving jerky actions in low-dimensional input by adding regularization terms \cite{siddharth2021caps}. 
Based on CAPS, Cao {\em et al.} proposed Image-based regularization for action smoothness (I-RAS) \cite{cao2023imagebased}, focusing on high-dimensional input scenarios.
Furthermore, based on Lipschitz constraints, recent contributions such as Locally Lipschitz Continuous Constraint (L2C2) \cite{kobayashi2022l2c2} and LipsNet \cite{lipsnet_2023} aim to ensure the policy function slowly changes to obtain the smoothness with small value of $K$-Lipschitz constants.

This paper proposes a novel approach to improve policy smoothness in DRL, named Gradient-based Conditioning for Policy Action Smoothness (Grad-CAPS). 
Grad-CAPS has several advantages compared to prior research.
Our method reduces the difference in the gradient of action, like the first-order derivative of the policy function, enhancing its ability to identify and smoothen zigzagging action sequences.
We also introduce displacement normalization to effectively regularize the action sequences regardless of the action scale across various scenarios and environments.
Our Grad-CAPS can serve as a new condition to other methods that based on Lipschitz constraints like CAPS, L2C2, or LipsNet to obtain smoothness behaviors.
As a regularization technique, our method can be incorporated into existing DRL algorithms.
Overall, our Grad-CAPS leads to smoother behaviors while maintaining comparable performance to other methods.

The main contributions of this paper are summarized as follows:
\begin{itemize}
    \item We propose Grad-CAPS, a method with regularization on the first-order derivative of the policy function, which effectively reduces zigzagging trajectories. 
    \item  We introduce displacement normalization to enable our method to adapt to invariant action scales and generalize across diverse environments.
    \item In our experiments, we demonstrate that integrated Grad-CAPS with different reinforcement learning algorithms clearly outperforms CAPS and other methods without CAPS on various robotic-related tasks while preserving both the expressiveness and smoothness of the policy.
\end{itemize}

\section{Backgrounds}

\subsection{Conditioning for Action Policy Smoothness (CAPS)}
\label{subsec:review_caps}
Mysore {\em et al.}\cite{siddharth2021caps} proposed a regularization technique to minimize the action change, resulting in a smoother trajectory. 
The method is constructed by two regularization terms:
\begin{inparaenum}[1)]
    \item temporal smoothness term, and 
    \item spatial smoothness term.
\end{inparaenum}

The temporal smoothness term $L_{temp}$ is the difference in action taken on two consecutive states $s_t$ and $s_{t+1}$.
Minimizing the term $L_{temp}$ is equivalently to reduce the difference between chronologically adjacent actions. 
The spatial smoothness term $L_{spat}$ is the difference between two actions taken on state $s$ and $s'$, where $s'$ is a similar state to $s$ and is generated by sampling from a normal distribution $\Phi$ centered around state $s$. 
Cao \textit{et al.} \cite{cao2023imagebased} used domain randomization to generate $s'$ when dealing with high-dimensional input. 
CAPS minimizes these two regularization terms while maximizing the original objective of reinforcement learning algorithm $J_\pi$ for a given policy $\pi$, formalized as follows.

Let ($\mathcal{A}$, $d_\mathcal{A}$) and ($\mathcal{S}$, $d_\mathcal{S}$) be two metric spaces, where both $\mathcal{A}$ and $\mathcal{S}$ are action and state spaces respectively.
The distance metric $d_\mathcal{A}$ for $\mathcal{A}$ is based on the Euclidean distance, namely, for two actions $a_1$ and $a_2$, 
\begin{equation}
    d_{\mathcal{A}}(a_1,a_2) = ||a_1 - a_2 ||_2.
\end{equation}
The distance metric $d_\mathcal{S}$ for $\mathcal{S}$ is based on the cardinal distance in an episode, namely, in a sequence of states, $\{s_1, s_2, ...\}$,
\begin{equation}
    d_{\mathcal{S}}(s_t,s_{t+k}) = k.
\end{equation}
Thus, two consecutive states  $d_{S}(s_t,s_{t+1})$ is simply one. 

An RL method decides an action for a given state $s$, based on a policy $\pi$. 
Namely, the action is taken by $a =\pi(s)$ in a deterministic manner, or by sampling on the distribution of a policy $a \sim \pi(\cdot | s)$ stochastically. 
The actions considered in the above regularization terms for CAPS are derived deterministically. 
Thus, $\pi(s)$ serves a mapping function from state $s$ to action $a$. 
The objective function of CAPS is to maximize $J_{\pi}^{\scalebox{0.5}{CAPS}}$ as follows:
\begin{equation}
J_{\pi}^{\scalebox{0.5}{CAPS}} = J_{\pi} - \lambda_{t}L_{temp} - \lambda_{s}L_{spat},
\label{equation:objective_functino_caps}
\end{equation}
\begin{equation}
L_{temp} = d_{\mathcal{A}}(a_{t},a_{t+1}), 
\label{equation:eq_temporal_loss}
\end{equation}
\begin{equation}
L_{spat} = d_{\mathcal{A}}(a_t,a'_t) \\
\end{equation}
\begin{equation*}
\text{where} \quad a'_t = \pi(s'_t)  \quad \text{and} \quad s'_t\sim \Phi(s_{t}).
\end{equation*}
The regularization weights $\lambda_{t}$ and $\lambda_{s}$ are used to control the impact of $L_{temp}$ and $L_{spat}$, respectively. 


The experiments for CAPS reported the significant influence of temporal smoothness on action smoothness compared to spatial smoothness. 
So, this paper focuses exclusively on temporal smoothness and omits spatial smoothness.

\subsection{Lipschitz Constraints}
\label{sec:review_Lipschitz}

Some of the previous works \cite{siddharth2021caps,cao2023imagebased,kobayashi2022l2c2,lipsnet_2023} also investigated to use the Lipschitz constraint to obtain the smoothness trajectory.
Let us review Lipschitz constraints as follows.

\begin{definition}(Lipschitz Constraints)
\label{definition:lipschtz_constraints}
Let $(X,d_X)$, $(Y,d_Y)$ be two metric spaces, where $d_X$ and $d_Y$ denote the distance metrics on sets X and Y, respectively.
A function $f$:  $X \rightarrow Y$ is called \textit{$K$-Lipschitz continuous} if there exists a real constant $K \ge 0$ such that, $\forall x_1,x_2 \in X$,
\begin{equation}
    d_Y (f(x_1),f(x_2) ) \le K  d_X( x_1,x_2), 
    \label{equation:global_lipschitz}
\end{equation}
\end{definition}

A smaller constant value for $K$ indicates a smoother function $f$.
While computing the exact values of Lipschitz constants is proven to be an NP-hard problem \cite{scaman2019lipschitz}, many studies utilize various regularization techniques to approximate the optimal value of Lipschitz constraints $K$, thereby achieving smoother functions.

Temporal smoothness in CAPS can also be viewed from Lipschitz constraints, where $(S,d_S)$ and $(A,d_A)$ correspond to $(X,d_X)$ and $(Y,d_Y)$ respectively, and $\pi$ serves as $f$. 
Suppose that temporal smoothness satisfies $K$-Lipschitz continuity. Thus, 
\begin{equation}
    L_{temp} = d_{\mathcal{A}}(a_t,a_{t+1} ) \le K  d_S(s_t,s_{t+1}).
    \label{equation:global_lipschitz_1}
\end{equation}
Since $d_{\mathcal{A}}(a_t,a_{t+1}) = ||a_t - a_{t+1} ||_2$ and $d_S(s_t,s_{t+1}) = 1$ as above, the following holds: $||a_t - a_{t+1} ||_2 \le K$. 

When CAPS minimizes the difference in action for smoothness, it also implicitly makes the Lipschitz constant $K$ smaller in general. 
However, a small $K$ causes the issue of \textit{over-smoothing}, namely, leading to a loss of expressiveness, as reported in the work \cite{kobayashi2022l2c2}.
For this issue, this paper proposes a regularization method over the first-order derivative of the policy function that obtains both smoothness and expressiveness of the action trajectory.

\section{Our Approach}
\label{sec:approach}
\subsection{Gradient-based Condition for Action Policy Smoothness (Grad-CAPS)}
\label{sec:grad-caps}

As discussed in \autoref{subsec:review_caps} above, CAPS made the policy excessively smooth by minimizing action changes. 
In terms of Lipschitz constraints, CAPS 
enforces the policy function $\pi$ with a small $K$, therefore losing the capability to handle situations requiring the agent to change actions with agility. 
For this issue, this paper proposes a new regularization approach called Gradient-based Condition for Action Smoothness (Grad-CAPS), which minimizes the differences of action changes, that is, the first-order derivative of action, instead of actions. 

Before discussing our method, we first define the first-order Lipschitz constraints over $f'$, the first-order derivative of the function $f$, as below:

\begin{definition}(First-order Lipschitz Constraints)
Let $(X,d_X)$, $(Y,d_Y)$ be two metric spaces, as in Definition \ref{definition:lipschtz_constraints}.
Let $f$ be a function: $X \rightarrow Y$, and $f'$ be a first-order derivative of $f$: $X \rightarrow \Delta Y$. 
Thus, $(\Delta Y, d_{\Delta Y})$ is a metric space, where $d_{\Delta Y}$ denotes the distance metric on $\Delta Y$. 
A function $f$ is called first-order $K$-Lipschitz continuous and $f'$ is $K$-Lipschitz continuous, if there is a constant $K \ge 0$ satisfying the following:
\begin{equation}
    d_{\Delta Y}(f'(x_1) , f'(x_2)) \le K d_X ( x_1 , x_2 ). 
    \label{equation:first_order_k_lipschtz}
\end{equation}
\end{definition}
As described above, we introduce a novel temporal smoothness approach by minimizing the change in the first-order derivative of action instead of the change in actions. 
As in \autoref{subsec:review_caps} where $\pi(s)$ is defined as a mapping function from $\mathcal{S}$ to $\mathcal{A}$, we define $\pi'(s)$ to be a mapping function from $\mathcal{S}$ to $\Delta \mathcal{A}$, the first-order derivative of action space $\mathcal{A}$, which are action changes between two consecutive time steps, namely, 
\begin{equation}
\pi'(s_t) = a_{t} - a_{t-1}. 
\end{equation}
Let $\Delta a_t$ denote $\pi'(s_t)$. 
Then, $\Delta a_t$ forms a space $\Delta \mathcal{A}$, and 
($\Delta \mathcal{A}$, $d_{\Delta \mathcal{A}}$) is a metric space, where $d_{\Delta \mathcal{A}}$ is the distance metric in the space $\Delta \mathcal{A}$ based on the Euclidean distance. 
Namely, for two action changes $\Delta{a}_{t1}$ and $\Delta {a}_{t2}$:
\begin{equation}
    d_{\Delta \mathcal{A}}(\Delta {a}_{t1}, \Delta {a}_{t2}) = ||\Delta {a}_{t1} - \Delta {a}_{t2} ||
\end{equation}
Instead, temporal smoothness loss in Grad-CAPS is defined as:
\begin{flalign}
\centering
  &\!\begin{aligned}
    \indent \indent \indent \indent L_{temp}  
            &  = d_{\Delta \mathcal{A}}( \Delta a_t , \Delta a_{t+1})  &\\
            & = || (a_{t} - a_{t-1}) - (a_{t+1} - a_{t}) ||_2
\label{equation:temporal_loss_2}
  \end{aligned}&
\end{flalign}
From the viewpoint of first-order Lipschitz constraint, suppose that Grad-CAPS temporal smoothness satisfies the first-order Lipschitz continuity. Thus,
\begin{equation}
    L_{temp} = d_{\Delta \mathcal{A}}(\Delta a_t , \Delta a_{t+1}) \le K  d_\mathcal{S}(s_t,s_{t+1})
    \label{equation:global_lipschitz_2}
\end{equation}
Since $d_\mathcal{S}(s_t,s_{t+1})$ is 1 and $d_{\Delta \mathcal{A}}( \Delta a_t , \Delta a_{t+1}) = || (a_{t} - a_{t-1}) - (a_{t+1} - a_{t}) ||_2$ from above, the formula now becomes:
\begin{equation}
    ||(a_t - a_{t-1}) - (a_{t+1} - a_{t}) ||_2 \le K
    \label{equal:grad_CAPS}
\end{equation}

\begin{figure}[t]
    \includegraphics[width=0.9 \columnwidth]{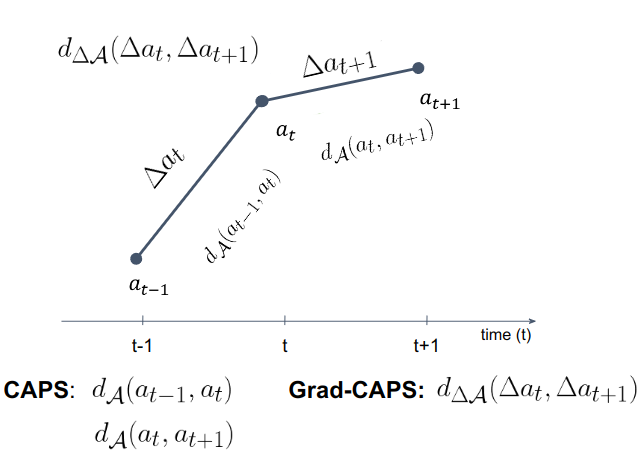}
    \caption{Differences of temporal smoothness loss between CAPS and Grad-CAPS.
    }
    \label{fig:difference_caps_gradcaps}
\end{figure}

\begin{figure}[h]
    \includegraphics[width=1.0 \columnwidth]{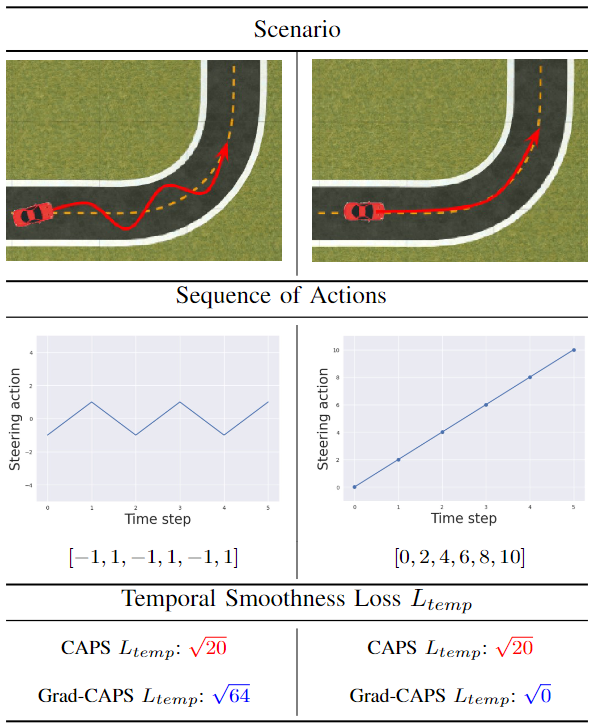}
    \caption{Two cases: one with a zigzagging sequence of actions (left) and the other with a sequence with constant action changes (right). The upper part shows corresponding car racing scenarios, and the lower part shows corresponding losses for CAPS and Grad-CAPS. 
    The cases show that CAPS fails to distinguish two sequences, while Grad-CAPS encourages stable action changes and penalizes zigzagging patterns.
    }
    \label{fig:zigzag_detection}
\end{figure}

In contrast to CAPS, which obtains smoothness by reducing the difference in action, our Grad-CAPS focuses on minimizing the difference in action change.
Fig. \ref{fig:difference_caps_gradcaps} illustrates the distinction between CAPS and Grad-CAPS.
Based on our definition, Grad-CAPS leads to improved regularization for distinguishing zigzagging sequences and larger effective action sequences compared to CAPS.
Fig. \ref{fig:zigzag_detection} illustrates the advantages of Grad-CAPS in recognizing zigzag trajectories over CAPS. 
We can observe that CAPS fails to distinguish between a zigzagging sequence and a sequence of actions with stable changes, whereas our Grad-CAPS allows for substantially stable changes in action while penalizing zigzagging patterns.
In short, CAPS defines a smooth trajectory as a slow change in action, while Grad-CAPS characterizes it as a stable change in action.

\subsection{Displacement Normalization}
Theoretically, our Grad-CAPS aims to reduce the difference in action changes as in (\ref{equation:temporal_loss_2}).

Nonetheless, optimizing the gradient of action change that satisfies \autoref{equal:grad_CAPS} still suffers the issue of over-smoothing of the policy and a loss of expressiveness, like CAPS, for the following reason. 

To minimize the term $L_{temp}$, the learning system tends to optimize the model in two ways. 
First, simply minimize the difference of consecutive $\Delta a_t$ and $\Delta a_{t+1}$, as we expected in \autoref{tab:example_displacement_definition}. 

Second, the system is distracted to minimize all $\Delta a_t$. 
For example, let $L_{temp}$ in (\ref{equation:temporal_loss_2}) satisfy the first-order $K$-Lipschitz constraints, as in \autoref{equation:global_lipschitz_2}. 
For achieving this, one way is simply to enforce to satisfy $K/2$-Lipschitz constraints, instead. 
Since $K/2$-Lipschitz constraints are satisfied, we have the following conditions:
\[ || a_{t} - a_{t-1}   ||_2 \le K/2, \; || a_{t+1} - a_{t}   ||_2 \le K/2 \]
Based on vector triangle inequality, obtain 

\begin{flalign}
\centering
  &\!\begin{aligned}
\indent \indent \indent L_{temp}    &= ||  (a_{t} - a_{t-1})  - (a_{t+1} - a_{t})  ||_2  \\
            &\le   || a_{t} - a_{t-1}   ||_2 + ||a_{t+1} - a_{t}||_2 &\\
            & \le K/2 + K/2 \le K.
  \end{aligned}&
\end{flalign}

That is, we can satisfy the first-order $K$-Lipschitz constraints by satisfying the $K/2$-Lipschitz constraints. 
Although this is less over-smoothing, the issue still exists.

To address this problem, we introduce displacement normalization to encourage the network to focus on optimizing the differences in the gradient of action instead of being distracted to optimize the differences in action. 

We first define the total action displacement $\delta_t$ around time $t$ as below:
\begin{equation}
     \delta_t = \Delta a_{t+1} + \Delta a_{t} =  (a_{t+1} - a_{t}) +  (a_{t} - a_{t-1}) = a_{t+1} - a_{t-1}
\end{equation}
Then, we divide the temporal smoothness loss in \autoref{equation:temporal_loss_2} by action displacement $\delta_{t}$:
\begin{equation}
     L_{temp} = d_{\Delta \mathcal{A}}( \frac{\Delta a_t}{\delta_t} , \frac{\Delta a_{t+1}}{\delta_t}) = \norm*{\frac{  \Delta a_t  - \Delta a_{t+1}}{  \delta_{t}}}_2
\end{equation}
Note that we add a small positive constant $\epsilon$ into denominator $D(=\delta_{t})$ to prevent from dividing zero, as follows. 
\begin{equation}
     L_{temp} = \norm*{\frac{  \Delta a_t  - \Delta a_{t+1} }{  \delta_{t} + \epsilon}}_2
\end{equation}
Besides, a small denominator $D$ also lets the whole term go to a huge number in practice. 
To solve this problem, we also apply the $tanh$ function to $1/D$ to limit the value range to $[-1,1]$ for stability. 
Thus our temporal regularization term in our Grad-CAPS becomes:
\begin{equation}
\centering
    L_{temp} = \norm*{\Delta a_t - \Delta a_{t+1}}_2 \tanh( \norm*{ \frac{1}{   (\delta_{t} + \epsilon)}
  }_2 )
\end{equation}

Normalizing the temporal loss function over the displacement enables many advantageous features in our proposed method.
First, Grad-CAPS regularizes the action sequence regardless of the action scale from different scenarios or environments.
Second, it magnifies the loss of the zigzagging pattern.
The low displacement in the zigzagging sequence magnifies the gradient loss by dividing the action displacement.
\autoref{tab:example_displacement_definition} shows examples of the benefit of adding displacement normalization.

\begin{table}[H]
\centering
\begin{subtable}{\linewidth}
\centering
\begin{tabular}{@{}c|c|c@{}}
\toprule

\thead{Action Sequence} 
& \thead{Grad-CAPS Loss \\ without normalization} 
& \thead{Grad-CAPS Loss \\ with normalization} \\ \midrule

\thead{\includegraphics[width=0.3 \textwidth]{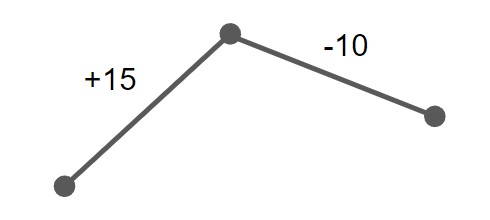}}
&\thead{$\sqrt{(15-(-10))^2}$ \\\\ $=\sqrt{625}$}
&\thead{$\sqrt{(\frac{15-(-10)}{5})^2}$ \\\\ $=\sqrt{25}$}\\ \midrule

\thead{\includegraphics[width=0.3 \textwidth]{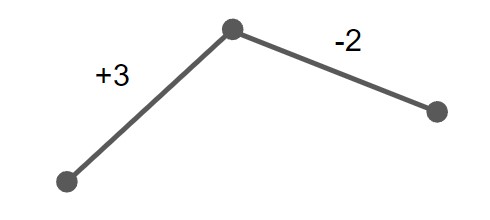}}
&  \thead{$\sqrt{(3-(-2))^2}$ \\\\ $=\sqrt{25}$}
&  \thead{$\sqrt{(\frac{3-(-2)}{1})^2}$ \\\\ $=\sqrt{25}$}\\ \midrule

\thead{\includegraphics[width=0.3 \textwidth]{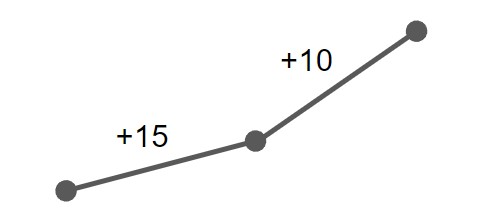}} 
&\thead{$\sqrt{(15-10)^2}$ \\\\ $=\sqrt{25}$}
&\thead{$\sqrt{(\frac{15-10}{25})^2}$ \\\\ $=\sqrt{\frac{1}{25}}$}\\ \midrule

\thead{\includegraphics[width=0.3 \textwidth]{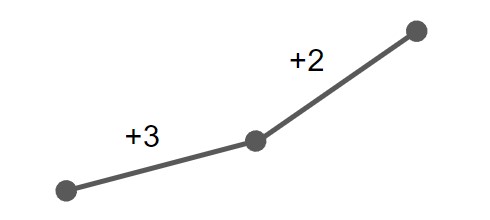}} 
&  \thead{$\sqrt{(3-2)^2}$ \\\\ $=\sqrt{1}$}
&  \thead{$\sqrt{(\frac{3-2}{5})^2}$ \\\\ $=\sqrt{\frac{1}{25}}$}\\ \midrule

\end{tabular}
\label{tab:toy_problem_sine_result}
\end{subtable}
\caption{Examples of sequences of actions for comparison of the temporal smoothness losses with and without normalization on different action scales.
        Grad-CAPS with displacement normalization magnifies the loss of the zigzagging pattern, demonstrating its ability to handle different action scales and penalizing the zigzagging pattern.}
\label{tab:example_displacement_definition}
\end{table}



\section{Experiment Results}

\subsection{Experiment Setup}
In the experiment, we compare the performance of the following agents:
\begin{inparaenum}[(a)]
\item Baseline agent, using Vanilla SAC algorithm. 
\item CAPS agent.
\item Grad-CAPS agent (ours), as described in \autoref{sec:approach}. 
\end{inparaenum}
These agents are tested across three tasks of varying difficulty levels: trajectory tracking in \autoref{sec:toy_example}, DeepMind Control Suite in \autoref{sec:dmcontrol}, and OpenAI Gym in \autoref{sec:openai_gym}.


To evaluate the performance of the agent, we use the following metrics:
\begin{inparaenum}[(a)]
    \item Average reward: the average accumulative reward achieved by an agent over 10 episodes, evaluated with the best policy.
    \item Action fluctuation: Similar to that in \cite{kobayashi2022l2c2,lipsnet_2023}, we also evaluate the smoothness of the policy with $L_2$ norm of temporal action change, $\big\| {a_t-a_{t-1}} \big\|_{2}$, namely, the average of all action changes.
    A smaller value of action fluctuation usually refers to smoother behavior.
    However, excessive smoothing does not always result in good performance.

\end{inparaenum}


\begin{figure}[]
    \centering
    \begin{subfigure}[b]{0.93\textwidth}
        \includegraphics[width=\textwidth]{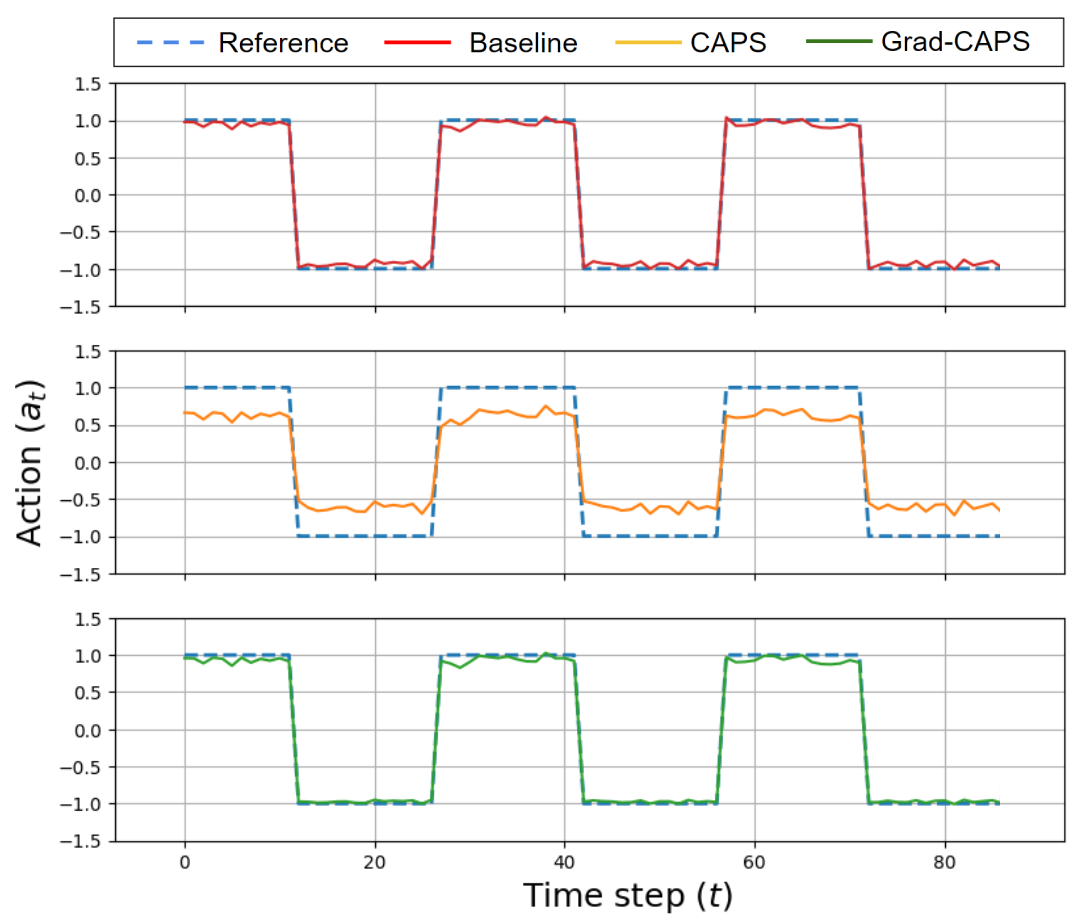}
        \caption{Square wave tracking experiment.}
        
        \label{fig:toy_example_1}
    \end{subfigure}
    \hfill
    \begin{subfigure}[b]{0.97\textwidth}
       \includegraphics[width=\textwidth]{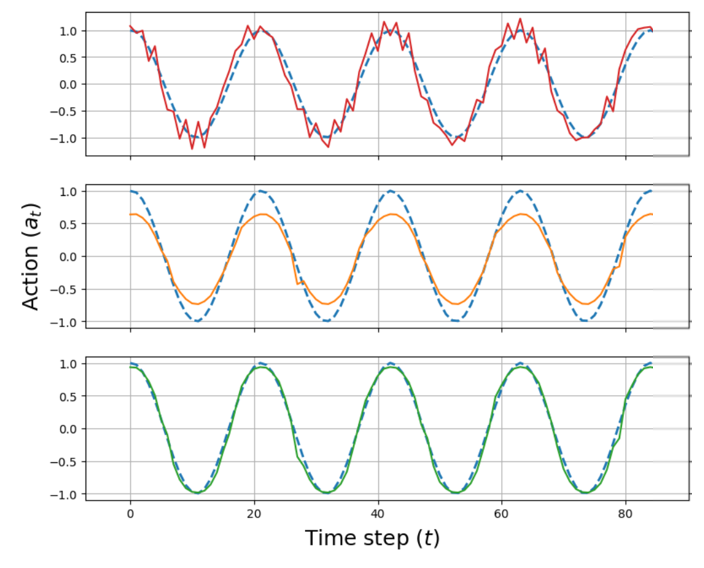}
        \caption{Cosine wave tracking experiment.}
        \label{fig:toy_example_2}
    \end{subfigure}

        
    
\caption{The referenced trajectories: (a) a square wave and (b) a cosine wave.
    The CAPS agent tends to over-smooth the action, leading to a loss of expressiveness in tracking the reference path.
    The Grad-CAPS agent performs better in following the reference path while maintaining smoothness.
    }
\label{fig:toy_example}

\end{figure}

\begin{table}[t]
\centering

\begin{subtable}{\linewidth}
\centering
\begin{tabular}{@{}lcc@{}}
\toprule
Method         & Square Wave $\uparrow$ & Cosine Wave $\uparrow$    \\ \midrule
Baseline       & -14.3 ± 0.6           & -16.6 ± 2.6               \\
CAPS           & -18.6 ± 0.7           & -20.7 ± 2.5              \\
Grad-CAPS  & \textbf{-14.3 ± 0.5}    & \textbf{-14.2 ± 1.8}      \\ \bottomrule
\end{tabular}
\caption{The average reward}
\label{tab:toy_problem_square_result}
\end{subtable}

\begin{subtable}{\linewidth}
\centering
\begin{tabular}{@{}lccc@{}}
\toprule
Method         & Square Wave $\downarrow$       & Cosine Wave $\downarrow$  \\ \midrule
Baseline       & 0.15 ± 0.4                    &  0.29 ± 0.19              \\
CAPS           & $\textbf{0.14 ± 0.2}^{\dag}$  &  $\textbf{0.09 ± 0.03}^{\dag}$   \\
Grad-CAPS      & 0.14 ± 0.4                    & {0.17 ± 0.12}             \\ \bottomrule
\end{tabular}
\caption{Action fluctuation.
${\dag}$ denotes the excessive smoothness that causes severe degradation in the agent's performance.}
\label{tab:toy_problem_sine_result}
\end{subtable}
\caption{Results of trajectory tracking shown in Fig. \ref{fig:toy_example}.}

\label{tab:compare_toy_example}
\end{table}

\begin{figure}[h]
     \centering   
     \begin{subfigure}[b]{0.49\textwidth}
         \centering
         \includegraphics[width=0.9 \textwidth]{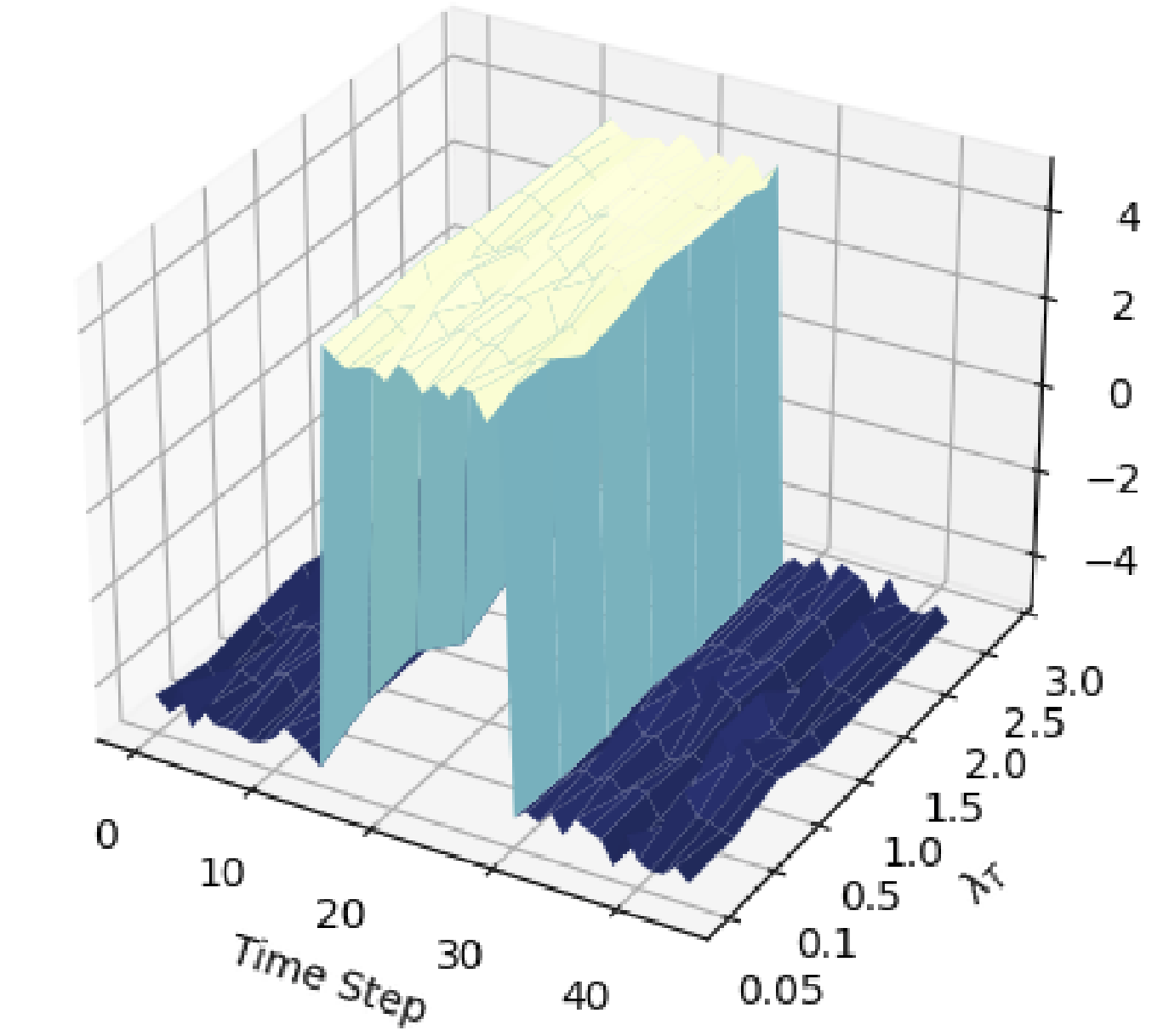}
         \label{}
     \end{subfigure}
     \begin{subfigure}[b]{0.49 \textwidth}
         \centering
         \includegraphics[width=0.9 \textwidth]{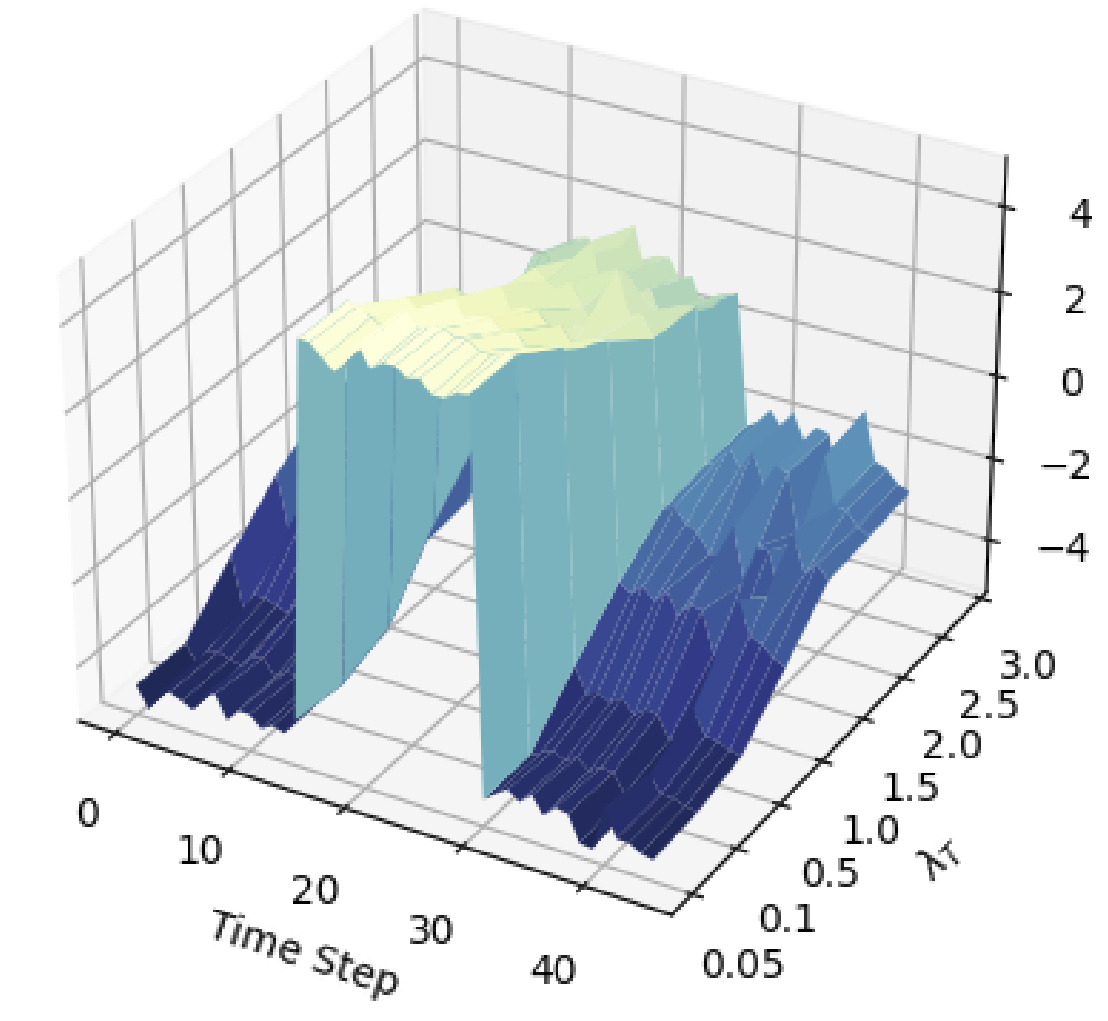}
         \label{}
     \end{subfigure}
     \begin{subfigure}[b]{0.49\textwidth}
         \centering
         \includegraphics[width=0.9 \textwidth]{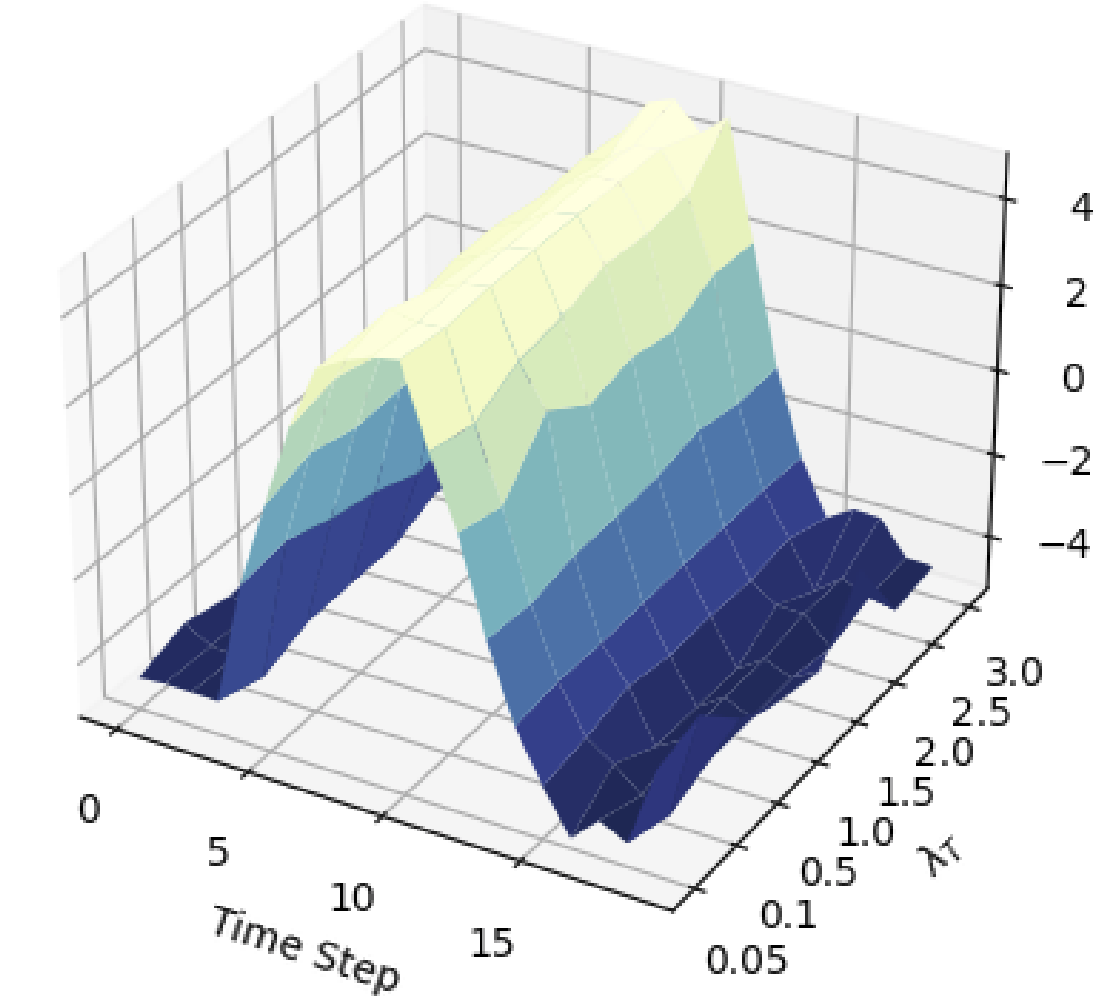}
         \caption{Grad-CAPS}
         \label{}
     \end{subfigure}
     \begin{subfigure}[b]{0.49 \textwidth}
         \centering
         \includegraphics[width=0.9\textwidth]{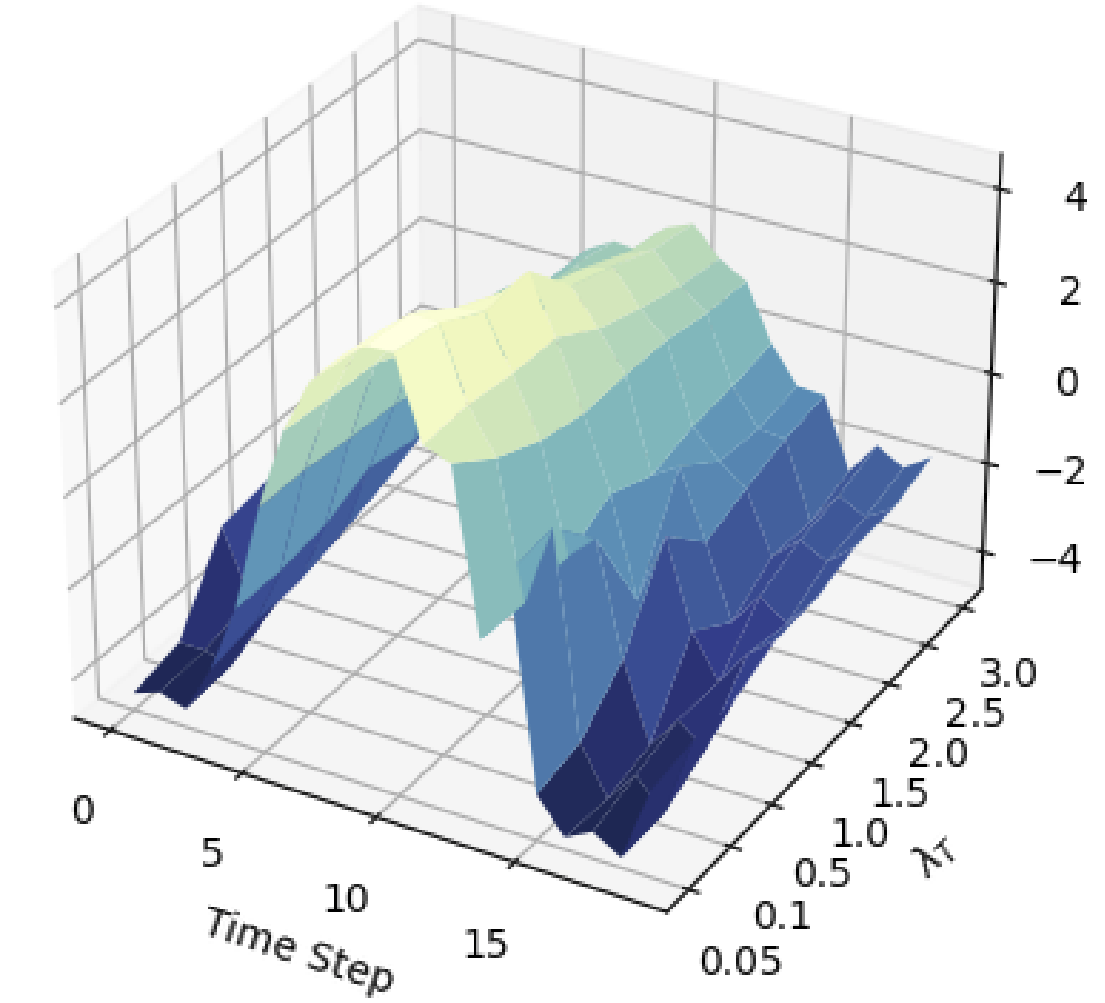}
         \caption{CAPS}
         \label{}
     \end{subfigure}
     \caption{Temporal loss weight $\lambda_T$ ablation study on wave tracking experiments.}
    \label{fig:lambda-experiment}
\end{figure}

\begin{table*}[ht!]
\centering
\begin{tabular}{lcccc} 
\hline
\multicolumn{1}{c}{\multirow{2}{*}{Algorithms}} & \multicolumn{4}{c}{Environments}                                                                    \\ 
\cline{2-5}
\multicolumn{1}{c}{}                           & Reacher           & Cartpole       & Walker             & Ball-In-Cup         \\
\hline
SAC - Vanilla                                    & 977.20 ± 10.30          & 856.45 ± 0.31          & 559.11 ± 15.00          & 976.49 ± 22.92          \\
SAC - CAPS                                       & 980.56 ± 11.87         & 807.50 ± 6.39          & 498.37 ± 86.77          & 977.55 ± 15.27           \\
SAC - Grad-CAPS                                  & \textbf{982.20 ± 11.40} & \textbf{856.60 ± 0.39} & \textbf{561.15 ± 12.97} & \textbf{979.79 ± 16.62}   \\ 
\hline
TD3 - Vanilla                                    & 936.94 ± 6.99\phantom{0}          & \textbf{860.27 ± 0.97} & 186.11 ± 20.33          & 978.35 ± 12.26          \\
TD3 - CAPS                                       & 909.34 ± 3.79\phantom{0}          & 849.45 ± 0.45          & 555.90 ± 12.59          & 978.18 ± 13.02          \\
TD3 - Grad-CAPS                                  & \textbf{985.10 ± 9.30\phantom{0}}         & 854.99 ± 0.32          & \textbf{663.10 ± 14.70} & \textbf{980.04 ± 12.38} \\ 
\hline
DDPG - Vanilla                                   & 865.14 ± 8.37\phantom{0}          & 877.74 ± 0.28          & {\phantom{0}34.91 ± 14.27}           & 977.78 ± 18.77           \\
DDPG - CAPS                                      & 928.82 ± 9.64\phantom{0}          & 857.94 ± 0.24          & 627.72 ± 18.10          & 974.94 ± 14.29         \\
DDPG - Grad-CAPS                                 & \textbf{928.26 ± 4.24\phantom{0}} & \textbf{880.13 ± 0.36} & \textbf{670.05 ± 19.20} & \textbf{980.21 ± 13.16}   \\
\hline
D4PG - Vanilla  & 970.10 ± 13.63 & 879.77 ± 0.65 & 752.97 ± 09.92 & 987.30 ± 13.12 
            \\
D4PG - CAPS & 982.00 ± 10.68 & 881.44 ± 0.73 & 748.59 ± 08.37 & \textbf{987.60 ± 10.36} 
\\
D4PG - Grad-CAPS  &  \textbf{989.10 ± 14.85} &
\textbf{882.49 ± 0.34} &
\textbf{779.87 ± 05.46} &
983.30 ± 15.49  

  \\
\hline
\end{tabular}
\caption{The average reward on DMControl.
The higher value of the average reward shows a better performance.
}
\label{tab:average_dm_control}
\end{table*}

\begin{table*}[ht]
\centering
\begin{tabular}{lcccc} 
\hline
\multicolumn{1}{c}{\multirow{2}{*}{Algorithms}} & \multicolumn{4}{c}{Environments}                                                              \\ 
\cline{2-5}
\multicolumn{1}{c}{}                            & Reacher           & Cartpole     & Walker            & Ball-In-Cup   \\ 
\hline
SAC - Vanilla                                     & 24.10 ± 2.90\phantom{0}            & 0.53 ± 0.03         & 20.70 ± 0.50           & 14.10 ± 1.71         \\
SAC - CAPS                                        & 6.50 ± 0.40            & 0.48 ± 0.03          & \textbf{2.74 ± 0.3}           & \phantom{0}1.30 ± 0.13          \\
SAC - Grad-CAPS                                   & \textbf{6.37 ± 0.28}   & \textbf{0.41 ± 0.03} & {\phantom{0}4.90 ± 0.21}  & \textbf{\phantom{0}0.60 ± 0.27}  \\ 
\hline
TD3 - Vanilla                                     & 67.38 ± 19.32          & \textbf{0.97 ± 0.46} & {\phantom{0}35.94 ± 20.23}         & 46.18 ± 7.99         \\
TD3 - CAPS                                        & 40.39 ± 19.03          & 1.61 ± 0.19          & 37.04 ± 0.73          & \textbf{39.29 ± 1.00}  
 \\
TD3 - Grad-CAPS                                   & \textbf{66.20 ± 22.15} & 2.14 ± 0.71          & \textbf{33.04 ± 0.74} & 43.59 ± 1.17      
   \\ 
\hline
DDPG - Vanilla                                    & 64.94 ± 19.11          & \textbf{1.34 ± 0.70} & {\phantom{0}33.82 ± 24.29}         & 39.56 ± 0.96          
 \\
DDPG - CAPS                                       & 95.79 ± 21.29          & 8.14 ± 0.38          & \textbf{31.20 ± 1.02} & 10.32 ± 7.71         
 \\
DDPG - Grad-CAPS                                  & \textbf{55.45 ± 22.62} & 2.13 ± 1.34          & 35.81 ± 0.90          & \textbf{\phantom{0}2.04 ± 4.46} 
 \\
\hline
D4PG - Vanilla &
45.47 ± 23.09 &
1.56 ± 0.02 &
27.55 ± 9.62 &
18.40 ± 0.74
\\
D4PG - CAPS &         
\textbf{32.17 ± 05.71} &
1.92 ± 0.05 &
\textbf{23.40 ± 5.61} &
18.93 ± 2.57
\\
D4PG - Grad-CAPS &
 46.14 ± 31.26 &
\textbf{1.25 ± 0.01} &
24.28 ± 3.74 &
\textbf{18.38 ± 1.31} 
\\
\hline
\end{tabular}
\caption{Action fluctuation on DMControl.
    The values are in units of $10^{-2}$ their original values.
    The lower value of action fluctuation shows a better smoothness trajectory.
    }
    
\label{tab:action_fluctuation_ratio_dm_control}
\end{table*}

\subsection{Trajectory Tracking}
\label{sec:toy_example}
In this section, we aim to assess the fundamental capability of the agent to generate precise action sequences or patterns, simulating the moving trajectory in general autonomous controlling tasks. The objective of this task is to follow the referenced trajectory accurately.

The trajectory tracking task is designed as follows.
The observation at time $t$ is the current position of the agent $pos_t$ and the generated target point $tar_{t}$.
From $pos_t$ and $tar_t$, we calculate the distance from the agent's position to the generated target point denoted as $dist_t$.
The state representation input to the agent is a $[2\times1]$ dimensional vector, defined as $s_t = [dist_t, pos_t]$.
For a given state representation $s_t$, the agent is required to predict the next target point $\overline{tar}_{t+1}$.
The reward is the negative of the difference between the predicted point  $\overline{tar}_{t+1}$ and the next generated target point $tar_{t+1}$, i.e., $-||\overline{tar}_{t+1} - tar_{t+1}||$.

We design two referenced trajectories: square and cosine waves, which serve as toy problems. 
The square wave goal is to evaluate the expressiveness of the policy, while the cosine wave is to test the smoothness of the policy.
Through these tests, we analyze the ability of agents to generate actions that align with specific patterns effectively.

\autoref{tab:compare_toy_example} compares the results of different agents for the toy examples.
From the table, Grad-CAPS clearly outperforms the other two in terms of average rewards. 
We observe that excessive regularization of action differences in the CAPS agent leads to a significant loss in policy expressiveness in both tests, though the agent has the least action fluctuation.
The baseline agent performs well in the square wave tracking task; however, in the cosine wave tracking task, the lack of smoothness regularization results in a zigzagging trajectory when following the designed trajectory.
In contrast, our regularization method outperforms other methods while effectively reducing zigzagging in actions and enabling the agent to stabilize action changes.

Fig. \ref{fig:toy_example} visualizes the predicted trajectory following the reference path of different agents.
From the observation of both square wave and cosine wave tracking tasks, Grad-CAPS performs well by following the reference path while maintaining a reasonably smooth trajectory.

     


Based on the above task for tracking square and cosine waves, we further investigate how the regularization weight $\lambda_{T}$ affects the performance of Grad-CAPS. 
(Note that the regularization weight $\lambda_{S}$ is ignored in this paper.)
In our experiment, we let $\lambda_{T}$ vary from 0.05 to 3.0.
Fig. \ref{fig:lambda-experiment} shows the results for different $\lambda_{T}$ by stacking the waves predicted by the agents in a 3D space.
From this figure, the performance of our Grad-CAPS agent in wave tracking remains consistent regardless of different weights. 
Thus, the Grad-CAPS agent demonstrates reliable tracking capabilities.
On the other hand, the CAPS agent exhibits distinct behaviors based on the weight settings. 
The CAPS agent performs better as weights become lower, but it produces more zigzagging patterns at the same time. 
On the other hand, higher weights result in a loss of expressiveness for the CAPS agent, indicating a decline in its ability to capture and represent the wave dynamics effectively.
Overall, the above highlights the robustness of our Grad-CAPS agent in wave tracking tasks, while it is sensitive for the CAPS agent to choose a temporal weight for the performance. 
In the rest of our experiments, we set the $\lambda_T = 1$ as in \cite{siddharth2021caps}.

\subsection{DeepMind Control Suite}
\label{sec:dmcontrol}
In this experiment, we evaluated the performance of our Grad-CAPS with a more complex environment, the DeepMind Control Suite \cite{tassa2018deepmind} (referred to as DMControl in this paper).
The DMControl provides a set of well-designed continuous control tasks involving interactions with diverse robotic systems, such as manipulating robotic arms or maintaining balance in dynamic scenarios.
Specifically, we selected four tasks from DMControl as follows:

For Cartpole, we choose Swingup, whose objective is to swing up and balance a pole attached to a cart.
For Reacher, we choose  Easy, whose objective is to control two joint motors of a robot arm to maneuver the endpoint toward a designated position without considering the complex dynamic influences present in other environments.
For Ball-In-Cup, we choose Catch, whose objective is to control an actuated planar receptacle to swing and catch a ball attached to its bottom.
For Walker, we choose Run, whose objective is to run forward as fast as possible, which requires the agent to make more rapid and aggressive action changes.

We compare four different reinforcement learning algorithms, including SAC \cite{Haarnoja2018SAC}, TD3 \cite{fujimoto2018td3}, DDPG \cite{lillicrap2019ddpg}, and D4PG \cite{barthmaron2018d4pg}, without smoothness terms (denoted as Vanilla) and with smoothness terms (denoted as CAPS and Grad-CAPS).

The results are presented in \autoref{tab:average_dm_control} and \autoref{tab:action_fluctuation_ratio_dm_control}, showing the average reward and action fluctuation of different agents in the DMControl environment.
The findings indicate that adding the smoothness term Grad-CAPS generally outperforms other configurations (Vanilla and CAPS) in terms of average rewards and smoothness values across different algorithms and tasks.
We also observe that the SAC agent has a better smoothness value compared to TD3, DDPG, and D4PG.
Grad-CAPS consistently demonstrates improved outcomes when combined with various reinforcement learning algorithms and tasks for controlling robot tasks, indicating its potential for enhancing the performance of the agent while maintaining comparable smoothness behaviors.
This showcases a promising approach for practical application in robotic scenarios where the smooth action trajectory of agents is essential.
Demonstration results are provided in the accompanying video of this paper.





    

\subsection{OpenAI Gym}
\label{sec:openai_gym}
In this experiment, we select three tasks from OpenAI Gym \cite{gymlibrary}: Half-Cheetah, Humanoid, and Car-Racing. 
These tasks are used to evaluate the performances of agents on challenging high-dimensional robotics control tasks and car racing in complex dynamic environments.
Note that in Car-Racing we only train and evaluate on a single map.

\begin{figure}[t]
\centering
    \begin{subfigure}[t]{0.99\columnwidth}
        \includegraphics[width=\columnwidth]{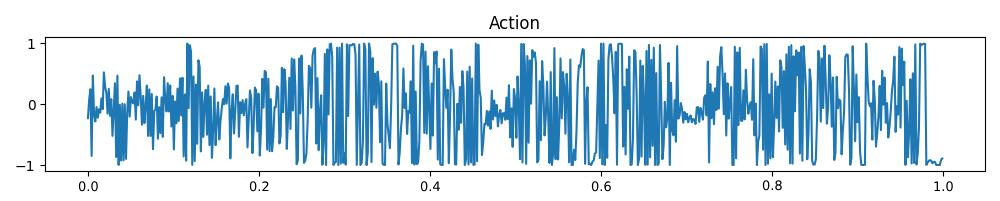}
        \caption{SAC - Vanilla.}
    \label{fig:vanila_sac_steering}
    \end{subfigure}
    \begin{subfigure}[t]{0.99\columnwidth}
        \includegraphics[width=\columnwidth]{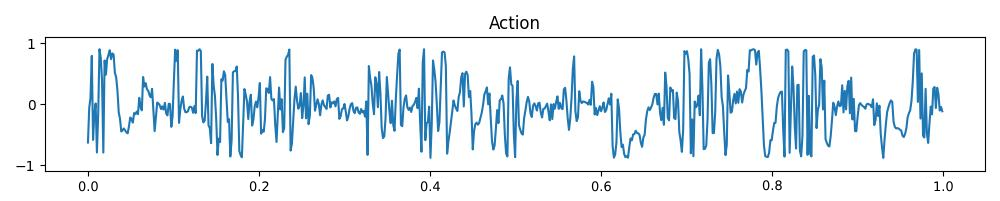}
        \caption{SAC - CAPS.}
        \label{fig:ras_steering_1}
    \end{subfigure}
    \begin{subfigure}[t]{0.99\columnwidth}
        \includegraphics[width=\columnwidth]{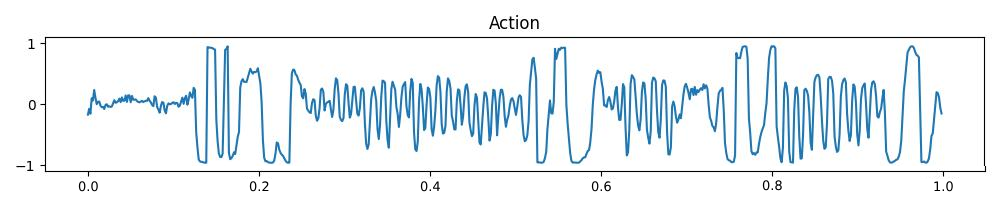}
        \caption{SAC - GradCAPS}
        \label{fig:ras_ir_steering}
    \end{subfigure}
\caption{Steering actions during a track of different agents in Car Racing environment. Grad-CAPS clearly obtains smoother steering action compared to other methods.
}
\label{fig:full_compare_steering}
\end{figure}

We implement and compare the SAC agents without and with smoothness terms.
The average rewards and smoothness values of different methods in the OpenAI Gym environment are reported in \autoref{tab:average_openai_control}.
We observe that CAPS significantly emphasizes smoothness, resulting in the lowest smoothness scores in Half-Cheetah and Humanoid.
However, focusing on smoothness reduces its ability to adapt to tasks requiring action changes, reducing overall performance in task completion compared to other methods.
In contrast, our Grad-CAPS agent still balances between performance and expressive action, therefore outperforming in all tasks while maintaining comparable smooth trajectories to CAPS.
Additionally, in Car-Racing, where achieving smoother actions is crucial for enhancing performance, our Grad-CAPS agent distinctly demonstrates better trajectory smoothness as in Fig. \ref{fig:full_compare_steering}.

\begin{table}[t]
\centering
\begin{subtable}{\linewidth}
\centering
\begin{tabular}{@{}lccc@{}}
\toprule
Method          & Half-Cheetah   & Humanoid & Car Racing  \\ \midrule
Vanilla               & 13060 ± 123               & 7530 ± 428  & 917            \\
CAPS                  & 10126 ± 15                & 7076 ± 212   & 932                     \\
Grad-CAPS & \textbf{13092 ± 97} & \textbf{8014 ± 390} & \textbf{942} \\ \bottomrule
\end{tabular}
\caption{The average reward on OpenAI Gym environments.
        The higher value of the average reward shows a better performance.
        }
\label{tab:average_reward_openai_control}
\end{subtable}

\begin{subtable}{\linewidth}
\centering
\begin{tabular}{@{}lccc@{}}
\toprule
Method          & Half-Cheetah      & Humanoid & Car-Racing   \\ \midrule
Vanilla           & 6.76 ± 0.02                   & 1.13 ± 0.02 & 0.35 \\
CAPS               & \textbf{4.11 ± 0.02}   & \textbf{0.69 ± 0.02}  & 0.15\\
Grad-CAPS          & 6.08 ± 0.02            & 0.75 ± 0.01 & \textbf{0.08} \\  \bottomrule
\end{tabular}
\caption{Action fluctuation on OpenAI Gym environments.
        The values are in units of $10^{-2}$. 
        The lower value of action fluctuation shows a better smooth trajectory.
        }
\label{tab:average_smoothness_openai_control}
\end{subtable}

\caption{Results of SAC agent with different regularization terms on OpenAI Gym environments.
}
\label{tab:average_openai_control}
\end{table}

\section{Conclusion}
In this paper, we addressed the critical issue of jerky trajectories in DRL and propose Grad-CAPS, a novel regularization method designed to reduce the problem of zigzagging actions.
Grad-CAPS allows the agent to expressively change the action to adapt to the environment while maintaining a smooth trajectory, balancing enhanced performance and smooth behavior execution.
We also introduced displacement normalization for adaptability to action scale, generalizing the use of our method across a wide range of applications.
Our experiments with DMControl and OpenAI Gym demonstrate that Grad-CAPS effectively enhances the performance of agents while maintaining a comparable level of smoothness compared to other methods across various reinforcement learning algorithms.
This improvement is beneficial for reinforcement learning techniques for robotic and autonomous systems where jerky actions can lead to inefficiencies or safety concerns.

\bibliographystyle{plain}
\bibliography{root}

\end{document}